\documentclass{article}
\usepackage[preprint]{spconf}
\usepackage{amsmath,graphicx,hyperref}
\usepackage{amssymb,enumitem,booktabs,multirow}
\usepackage{pifont}
\newcommand{\cmark}{\ding{51}} % ✓
\newcommand{\xmark}{\ding{55}} % ✗

\title{SFN-YOLO: Towards Free-Range Poultry Detection via Scale-aware Fusion Networks}
\name{
\parbox{\textwidth}{
\centering
Jie Chen$^1$ \quad
Yuhong Feng$^1$\sthanks{Corresponding author.} \quad
Tao Dai$^1$ \quad
Hao Wang$^2$ \quad
Hongtao Chen$^1$ \\
Zhaoxi He$^1$ \quad
Mingzhe Liu$^1$ \quad
Jiancong Bai$^1$
}
}

\address{$^1$ Shenzhen University, Shenzhen, China\\
$^2$ The Hong Kong University of Science and Technology (Guangzhou), Guangzhou, China}

\begin{document}
\ninept
\maketitle

\begin{abstract}

% Poultry detection in free-range environments remains challenging due to multiscale targets, occlusions, cluttered, and dynamic backgrounds, which often force trade-offs between accuracy and speed in existing detectors. We propose SFN-YOLO, a scale-aware fusion network that integrates local fine-grained features with global context to enhance perceptual capability in complex environments. Trained and evaluated on {\em M-SCOPE}, a new expansive dataset specifically constructed for diverse free-range scenarios, our model achieves 80.7\% mAP with only 7.2M parameters (35.1\% fewer than the baseline) while maintaining strong cross-domain generalization. The efficient real-time detection of SFN-YOLO can enable automated smart poultry farming. The code and dataset are available at https://github.com/chenjessiee/SFN-YOLO.

%%% by Tao Dai
Detecting and localizing poultry is essential for advancing smart poultry farming. Despite the progress of detection-centric methods, challenges persist in free-range settings due to multiscale targets, obstructions, and complex or dynamic backgrounds. To tackle these challenges, we introduce an innovative poultry detection approach named SFN-YOLO that utilizes scale-aware fusion. This approach combines detailed local features with broader global context to improve detection in intricate environments. Furthermore, we have developed a new expansive dataset ({\em M-SCOPE}) tailored for varied free-range conditions. Comprehensive experiments demonstrate our model achieves an mAP of 80.7\% with just 7.2M parameters, which is 35.1\% fewer than the benchmark, while retaining strong generalization capability across different domains. The efficient and real-time detection capabilities of SFN-YOLO support automated smart poultry farming.

\end{abstract}
\section{Introduction}
\label{sec:intro}

Object detection, a fundamental computer vision technique for identifying and localizing objects, is crucial to AI-driven smart poultry farming. It enables critical tasks, including flock counting, behavior analysis, and abnormality detection~\cite{Ambafi2025IoTPoultryReview,Dhattarwal2021SmartPoultry}, thus reducing labor costs and significantly boosting operational efficiency. 

The deployment of poultry detection in real-world farming environments presents a critical dilemma: high-accuracy models sacrifice inference speed, while lightweight models compromise precision. This inherent trade-off between speed and accuracy poses a significant challenge to mainstream detectors such as Faster R-CNN~\cite{ren2015faster}, Retinanet\cite{lin2017focal}, DETR~\cite{carion2020end}, RTDETR~\cite{zhao2024detrs}, and the YOLO series~\cite{yolov5,yolov8_ultralytics,yolo11_ultralytics,wang2025mamba,chen2025caf,xiao2025fbrt,ge2021yolox}, hindering their direct deployment. To overcome these challenges, recent research adapted existing architectures for poultry farming scenarios: EMSC-DETR~\cite{li2024efficient} enhances RT-DETR to manage multiscale and occlusion problems better; ChickTrack~\cite{2022ChickTrack} integrates YOLOv5 with a Kalman filter to achieve stable individual tracking; and ODBO~\cite{2025Only} employs a connector module to link YOLOv8-based behavior detection with identity recognition.

However, their performance significantly degrades the detection of small or obscured objects against complex and cluttered backgrounds~\cite{li2025survey, verma2023review}. This limitation arises as most of them are designed for controlled laboratory settings~\cite{wang2020real} or caged settings~\cite{zhuang2019detection} with enclosed scenes and simple backgrounds~\cite{yang2022deep}, are trained on limited datasets derived from such controlled environments. Consequently, the multiscale, cluttered, occluded, and dynamic nature of free-range farming environments demands a model that excels not only in balancing accuracy and speed, but also in cross-scene generalization~\cite{li2024efficient}.

In this paper, we propose {\em SFN-YOLO}, an efficient detection network for free-range poultry based on the YOLOv8 architecture. The core of our model is a novel {\em Scale-aware Fusion Module (SFM)}, designed explicitly to refine feature representations against key challenges such as scale variation, occlusion, and background clutter. The SFM employs a parallel structure: a local branch that captures fine-grained features for accurate localization, and a global branch that models contextual relationships to reduce ambiguities in cluttered or occluded scenes. A bidirectional guidance mechanism facilitates synergistic interaction between these two branches, enhancing representational coherence. To validate our approach and ensure its generalization across the dynamic conditions of free-range farms, we construct an expansive dataset, {\em M-SCOPE}, covering diverse poultry farming scenarios. Extensive experimental results and ablation studies demonstrate that SFN-YOLO achieves state-of-the-art (SOTA) performance and exhibits strong cross-domain generalization capability. In conclusion, the main contributions of this work are threefold:
\begin{itemize}[left=0em]
    \item We propose SFN-YOLO, an efficient scale-aware poultry detection network, and introduce a new expansive dataset comprising 6,600 images collected from diverse free-range scenarios.
    \item We design a scale-aware fusion module (SFM), which uses a bidirectional guidance mechanism between global context and local detail branches, significantly enhancing detection performance under challenges such as scale variation, occlusion, and cluttered backgrounds.
    \item Extensive experiments show that SFN-YOLO achieves SOTA performance, outperforming existing methods in cross-domain generalization while using 35.1\% fewer parameters than the baseline model, demonstrating its efficiency and generalization.
\end{itemize}

\section{Method}
\label{sec:format}

\begin{figure*}[!htb]
\centering
\includegraphics[width=\textwidth]{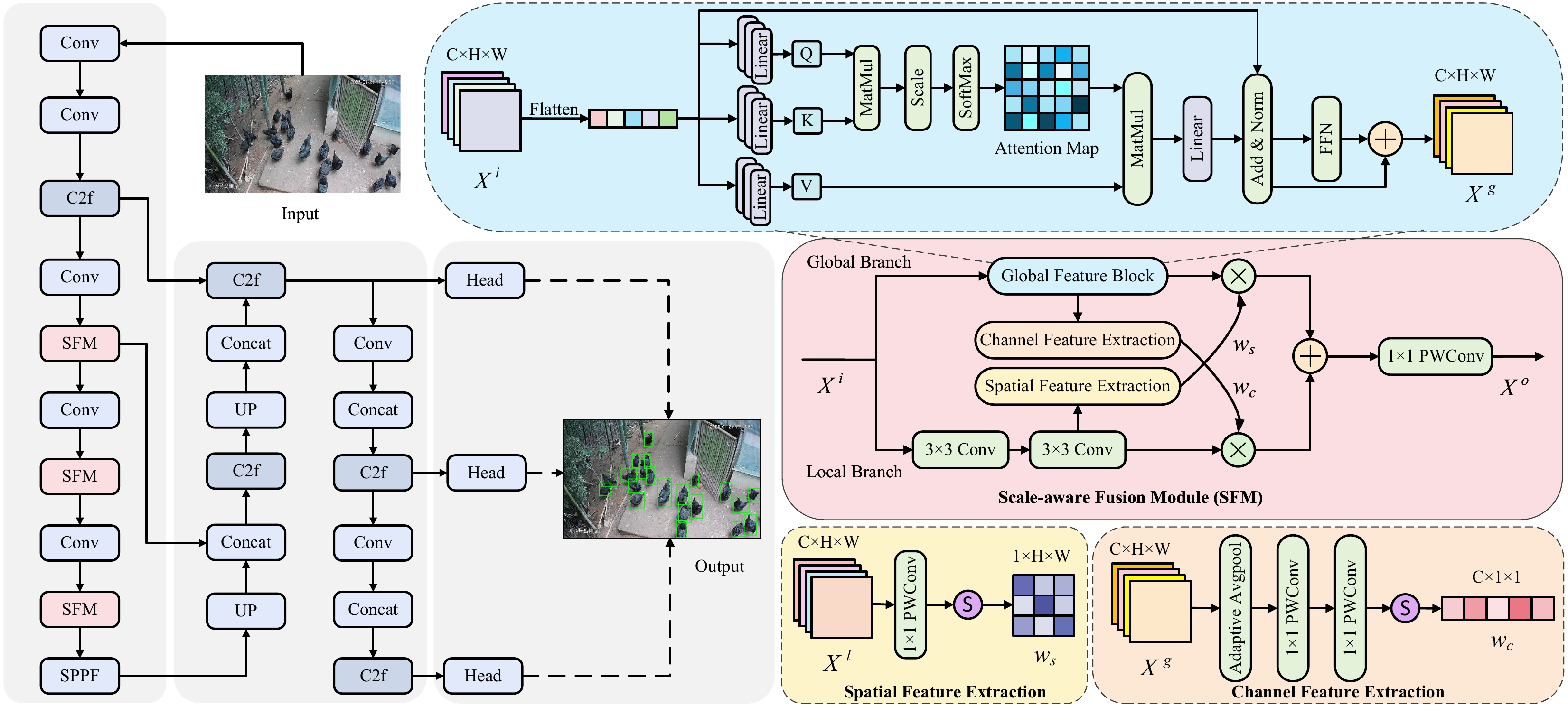} 
\caption{Framework of SFN-YOLO. The proposed SFM is embedded at three backbone stages. It implements a parallel local-global bidirectional guidance mechanism to enhance multiscale target perception by integrating fine-grained details with contextual information. Operations: 
\textcircled{\scriptsize S} (Sigmoid), 
\textcircled{+} (Element-wise Addition), and
\textcircled{\scriptsize $\times$} (Element-wise Multiplication).}
\label{fig:pic}
\vspace{-10pt}
\end{figure*}

\subsection{The SFN-YOLO Architecture}
Fig.~\ref{fig:pic} illustrates the overall architecture of SFN-YOLO, which introduces a novel SFM to enhance the backbone feature representation through the fusion of fine-grained local details and global contextual information. The SFM replaces the original cross stage partial with 2 convolutions-fusion (C2f) blocks to overcome their limited capacity in capturing long-range dependencies. Integrated at three backbone stages prior to the neck, this pre-fusion mechanism strengthens multiscale features with improved local precision and global context awareness, thereby boosting model's generalization in dynamic and cluttered backgrounds and under significant scale variation.

\subsection{The Scale-aware Fusion Module (SFM)}
The SFM integrates the local feature extraction capability of convolutional operations with the global contextual modeling ability of self-attention mechanisms. 
% It employs parallel local convolution and global self-attention branches to simultaneously capture local fine-grained details and global semantic context.
It employs parallel local convolution and global self-attention branches to address scale variation within intricate environments by simultaneously capturing local fine-grained details for small targets and global semantic context for large ones.
A bidirectional guidance mechanism further enhances representation learning: the global branch provides semantic priors for channel feature refinement, while the local branch corrects spatial prediction biases using fine-grained localization. This structured interaction improves feature discriminability and model generalization. As illustrated in Fig.~\ref{fig:pic}, the SFM is built on three core design principles: (1) Parallel feature extraction; (2) Bidirectional guidance; and (3) Feature fusion.

\noindent \textbf{Parallel Feature Extraction.} The SFM processes an input feature map $X^{i} \in \mathbb{R}^{C \times H \times W}$ through two parallel branches.
 
\textit{Local Convolutional Branch:} This branch extracts fine-grained local details, e.g. edges and textures, which are essential for identifying small-scale targets and accurately localizing the boundaries of larger or occluded objects. It consists of two \(3 \times 3\) convolutional layers, each followed by batch normalization and a SiLU activation. We formalize this operation as \(\mathcal{C}_{3 \times 3}(\cdot)\), yielding the local feature map: \(X^{l} = \mathcal{C}_{3 \times 3}(X^{i})\).

\textit{Global Self-Attention Branch:} This branch captures long-range dependencies to establish a global contextual representation, which aids in integrating features of large objects, reconstructing the shapes of occluded objects, and leveraging contextual cues to distinguish small objects from cluttered backgrounds. The global feature map \(X^g \in \mathbb{R}^{C \times H \times W}\) is generated by a Global Feature Block that performs a three-stage process: Reshaping, Self-Attention, and Feed-Forward Network.
\begin{itemize}[left=0em]
     \item \textit{Reshaping:} The input tensor \(X^{i}\) is reshaped into a sequence of tokens \(X^{f} \in \mathbb{R}^{N \times C}\), where \(N = H \times W\).
     \item \textit{Self-Attention:} The sequence \(X^{f}\) is fed into a multi-head self-attention (MHSA) module, in which query (\(Q\)), key (\(K\)), and value (\(V\)) tensors are derived through linear projections. To stabilize training, we apply L2 Normalization, denoted as \(\mathcal{N}(\cdot)\), to \(Q\) and \(K\), effectively converting their dot product into a cosine similarity measure. Furthermore, we introduce a learnable temperature parameter \(\gamma\) in place of the conventional fixed scaling factor, enabling dynamic adjustment of the attention distribution's sharpness. The operation within each attention head is formulated as shown in Eq.~(\ref{eq:attention}):
\begin{equation}
    \text{Attention}(Q, K, V) = \text{Softmax}\left(\frac{\mathcal{N}(Q)\mathcal{N}(K)^T}{\gamma}\right)V
    \label{eq:attention}
\end{equation}
We define the composite function \(\mathcal{A}\) as the sequential application of Layer Normalization and MHSA. The operation is expressed as: \(X'_{g} = X^{f} + \mathcal{A}(X^{f})\).

    \item \textit{Feed-Forward Network:} The updated sequence \(X'_g\) is then processed by a position-wise feed-forward network (FFN), which is formulated as a residual connection: \(X_g = X'_{g} + \mathcal{F}(X'_{g})\), where \(\mathcal{F}(\cdot)\) denotes the operation of layer normalization followed by the FFN. The output of the global branch, i.e., \(X^g\), is finally obtained by reshaping \(X_g\) back to its original 2D topology.
\end{itemize}

\noindent \textbf{Bidirectional Guidance.} Bidirectional attention guidance enables interactive enhancement between local and global features across both spatial and channel dimensions. This mechanism consists of two core components: spatial feature extraction and channel feature extraction.

\textit{Spatial Feature Extraction:} To enhance spatially significant regions for the global branch and refine spatial prediction accuracy using fine-grained details, a spatial attention map is constructed from local features \(X^{l}\). This is achieved via a \(1 \times 1\) convolution followed by a sigmoid activation, which reduces the channel dimension to one and produces a spatial weighting tensor \( w_{s} \in \mathbb{R}^{1 \times H \times W} \). The operation is formally defined as: \( w_{s} = \mathcal{S}(\mathcal{C}_{1\times1}(X^{l})) \), where \( \mathcal{S} \) denotes the sigmoid function and \( \mathcal{C}_{1\times1} \) represents the pointwise \( 1 \times 1 \) convolution.

\textit{Channel Feature Extraction:} The global branch produces channel-wise attention to semantically guide the local branch, following a mechanism similar to the Squeeze-and-Excitation (SE) block~\cite{se}. This process improves feature selection by emphasizing informative channels and suppressing less relevant ones using high-level semantic cues. A channel descriptor \(\mathbf{z} \in \mathbb{R}^{C}\) is first obtained via global average pooling over the spatial dimensions of the global feature map. The \(k\)-th element of \(\mathbf{z}\) can be calculated using Eq.~(\ref{c}):
\begin{equation}
z_k = \frac{1}{H \times W} \sum_{i=1}^{H} \sum_{j=1}^{W} X^{\text{g}, k}(i, j) \quad \text{for } k=1, \dots, C
\label{c}
\end{equation}
where \( X^{\text{g}, k}(i, j) \) denotes the value at location \( (i, j) \) in the \( k \)-th channel. The descriptor \(\mathbf{z}\) is then transformed through a two-layer MLP to capture channelwise dependencies. The resulting channel attention weight \( w_{c} \in \mathbb{R}^{C \times 1 \times 1} \) is computed as: \( w_c = \mathcal{S}(\mathcal{C}_{g} (\mathbf{z})) \), where \( \mathcal{C}_{g} \) denotes a mapping function implemented with two \( 1 \times 1 \) convolutional layers and a GELU activation.

\noindent \textbf{Feature Fusion.}  
The guidance maps are applied to the feature maps from the other branch via element-wise multiplication (with broadcasting for mismatched dimensions). This recalibrates local features using global context and focuses global features with local cues. The local feature \(X^{l}\) is guided as \(F_{l}\) (Eq.~\ref{eq:fl}), and the global feature \(X^{g}\) is guided as \(F_{g}\) (Eq.~\ref{eq:fg}). Finally, \(F_{l}\) and \(F_{g}\) are fused by element-wise addition, followed by a pointwise \(1 \times 1\) convolution to integrate the information, resulting in the final output \(X^{o}\) (Eq.~\ref{eq:final}). Residual connections stabilize training and ensure efficient information flow.

\begin{equation}
F_{l} = w_c  \ \textcircled{$\times$} \ X^{l}
\label{eq:fl}
\end{equation}
\begin{equation}
F_{g} = w_s \ \textcircled{$\times$} \ X^{g}
\label{eq:fg}
\end{equation}
\begin{equation}
X^{o} = X^{i} \ \textcircled{+} \ \mathcal{ C}_{ 1\times1} ( F_{l} \ \textcircled{+} \ F_{g})
\label{eq:final}
\end{equation}
where \textcircled{$\times$} is element-wise multiplication, \textcircled{+} is element-wise addition. The SFM adopts the YOLOv8 loss function, integrating CIoU, BCE, and DFL to balance localization, classification, and regression.

\section{Experiments}
\noindent\textbf{M\mbox{-}SCOPE Dataset.} 
We introduce the M\mbox{-}SCOPE (Multiscale Cluttered \& Occluded Poultry Environment) dataset, designed for object detection in cluttered free-range environments, with annotations for the ``chicken'' class. The dataset includes videos captured across 14 scenes, covering both indoor scenes and outdoor free\mbox{-}range scenes. Outdoor scenes were recorded from January to April 2025 at a resolution of \(2560\times1440\) and 24\,fps. All annotations were produced on the T\mbox{-}Rex Label platform~\cite{trex} following the VOC2011 standard~\cite{voc}. To facilitate the study of generalization under distribution shift, we partitioned the dataset into two subsets using a closed-set protocol. The in-domain (ID) dataset, used for training, validation, and testing, consists of 10 outdoor free-range scenes, while the out-of-domain (OOD) dataset, composed of 4 novel scenes, is reserved exclusively for testing. A statistical overview of these splits is in Table~\ref{t1}.

\noindent\textbf{Implementation Details. }All experiments were conducted on the M\mbox{-}SCOPE using an NVIDIA GeForce RTX 3090 GPU. The models were implemented in PyTorch with input images resized to 640$\times$640. Training was performed for 300 epochs with the stochastic gradient descent (SGD) optimizer, using a momentum of 0.937, weight decay of 0.0005, batch size of 2, an initial learning rate of 0.01, and an early stopping patience of 50 epochs. The number of heads in the self-attention mechanism was set to 8.

\begin{figure*}[htb]
\centering
\includegraphics[width=\textwidth]{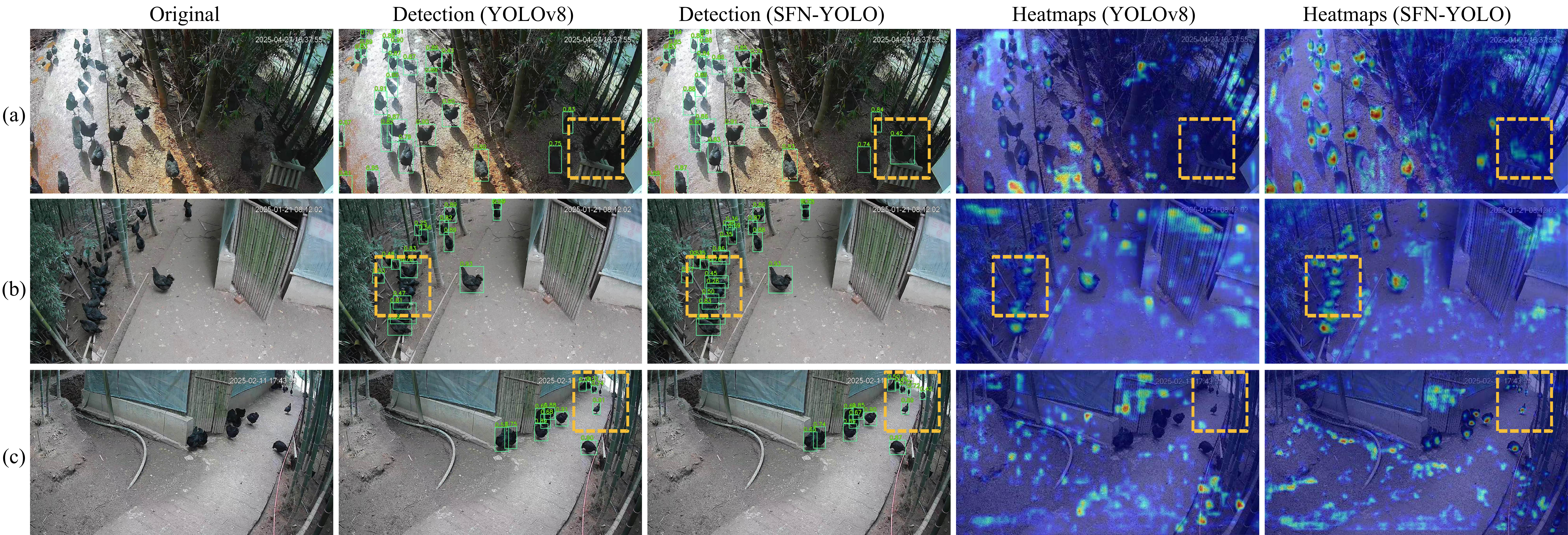} 
\caption{Qualitative comparison of YOLOv8 and SFN-YOLO on M\mbox{-}SCOPE: Detection visualization and heatmaps demonstrate the enhanced focus and accuracy of our model in challenging environments.}
\label{fig:vis}
\vspace{-15pt}
\end{figure*}

\noindent\textbf{Quantitative Result. }To evaluate the effectiveness of our proposed SFN-YOLO, we conducted comparisons with several SOTA object detectors on M\mbox{-}SCOPE. As shown in Table~\ref{tc}, our models achieve a compelling balance between accuracy and efficiency. The SFN-YOLO model achieves the highest mAP of 80.7\%, outperforming all competitors. Compared to the baseline YOLOv8s, it improves mAP by 0.4\% , reduces parameters by 35.1\% (from 11.1M to 7.2M) , and achieves a real-time speed of 112.4 FPS.
Additionally, it achieves an impressive AP75 score of 90.0\%, indicating superior localization accuracy. In the lightweight category, the SFN-YOLOn model achieves an mAP of 78.8\%, surpassing models like YOLOv8n (78.1\%) and YOLOv5n (77.8\%). It is the most parameter-efficient model, with just 2.0M parameters, 33.3\% fewer than YOLOv8n. SFN-YOLOn also shows improvements in small object detection, with an APS of 74.9\%, a 2.5\% gain over YOLOv8n (72.4\%) and a 3.2\% gain over YOLOv5n (71.7\%). These results confirm that both SFN-YOLO variants set a new SOTA on the M\mbox{-}SCOPE, offering superior accuracy with fewer parameters than existing detectors.

%\caption{Dataset Statistics. Box sizes are categorized as Small (S), Medium (M), and Large (L).}

\begin{table}[t]
\centering
\caption{Dataset Statistics with Box Size Categorized: Small (S), Medium (M), and Large (L).}
\label{t1}
\renewcommand{\arraystretch}{1} 
\footnotesize    
\setlength{\tabcolsep}{5pt} 
\begin{tabular}{c|c|cc|ccc} 
\toprule
\multirow{2}{*}{\textbf{Dataset}} & \multirow{2}{*}{\textbf{Scenes}} & \multirow{2}{*}{\textbf{Images}} & \multirow{2}{*}{\textbf{Boxes}} & \multicolumn{3}{c}{\textbf{Box Size(\%)}} \\ 
\cline{5-7}
 & & & & \rule{0pt}{1.2em}\textbf{S} & \textbf{M} & \textbf{L} \\

\midrule
\multirow{3}{*}{ID} & train & 4000 & 68,850 & 4.20 & 37.54 & 58.26 \\
& val & 500 & 8,642 & 2.88 & 37.87 & 59.25 \\
& test & 500 & 8,932 & 4.29 & 38.15 & 57.56 \\ 
\midrule
\multirow{4}{*}{OOD} & outdoor1 & 500 & 7,845 & 0.04 & 30.04 & 69.92 \\
& outdoor2 & 500 & 4,076 & 0.00 & 13.69 & 86.31 \\
& indoor & 500 & 8,327 & 6.14 & 22.43 & 71.43 \\
& snow & 100 & 386 & 0.52 & 14.25 & 85.23 \\ 
\bottomrule
\end{tabular}
\vspace{-15pt}
\end{table}

%\caption{Comparison of AP(\%) and Params(M) on M\mbox{-}SCOPE by using our methods with different real-time object detectors.}

\begin{table}[t]
\centering
% \caption{Accuracy (AP) vs. Efficiency (Params) on M\mbox{-}SCOPE: Real-Time Detectors with Our Method.}
\caption{Accuracy (AP, \%), Efficiency (Params, M), and Real-time (FPS) on M\mbox{-}SCOPE: A Comparison of Object Detectors.}
\renewcommand{\arraystretch}{1} 
\footnotesize 
\setlength{\tabcolsep}{2pt} 
\begin{tabular}{c|cccc|c|c} 
\toprule
\textbf{Model} & \textbf{mAP$\uparrow$} & \textbf{AP50$\uparrow$} & \textbf{AP75$\uparrow$} & \textbf{APS$\uparrow$} & \textbf{FPS$\uparrow$} & \textbf{Params$\downarrow$}
 \\
\midrule
Faster-RCNN\cite{ren2015faster} & 63.2 & 90.2 & 74.4 & 57.1 & 34.4 & 41.3 \\
\midrule
Retinanet\cite{lin2017focal} & 59.4 & 88.7 & 66.2 & 46.1 & 41.6 & 19.8 \\
\midrule
YOLOx\cite{ge2021yolox} & 68.2 & 94.0 & 79.3 & 60.3 & 44.2 & 8.9 \\
\midrule
YOLOv5n\cite{yolov5} & 77.8 & 96.1 & 87.1 & 71.7 & 232.6 & \underline{2.5} \\
YOLOv5s\cite{yolov5} & 80.1 & 96.4 & 88.7 & \underline{75.4} & 238.1 & 9.1 \\
\midrule
YOLOv8n\cite{yolov8_ultralytics} & 78.1 & 96.1 & 87.4 & 72.4 & \textbf{270.3} & 3.0 \\
YOLOv8s\cite{yolov8_ultralytics} & \underline{80.3} & 96.3 & 88.6 & 74.9 &  \underline{243.9} & 11.1 \\
\midrule
YOLO11n\cite{yolo11_ultralytics} & 77.7 & 96.0 & 87.0 & 71.9 & 208.3 & 2.6 \\
YOLO11s\cite{yolo11_ultralytics} & 80.1 & 96.3 & 88.5 & \underline{75.4} & 185.2  & 9.4 \\
\midrule
RT-DETR-R50\cite{zhao2024detrs} & 71.2 & 94.4 & 82.4 & 61.0 & 49.8 & 41.9 \\
RT-DETR-R101\cite{zhao2024detrs} & 71.4 & 94.3 & 81.9 & 60.5 & 43.7 & 60.9 \\
\midrule
MambaYOLO-T\cite{wang2025mamba} & 61.3 & 90.7 & 71.2 & 49.3 & 84.7 & 6.0 \\
MambaYOLO-B \cite{wang2025mamba}& 65.0 & 92.3 & 75.3 & 53.7 & 81.3 & 21.8 \\
\midrule
SFN-YOLOn (Ours) & 78.8 & \underline{96.7} & \underline{88.9} & 74.9 & 116.3 & \textbf{2.0} \\
SFN-YOLOs (Ours) &\textbf{80.7} & \textbf{96.8} & \textbf{90.0} & \textbf{75.8} & 112.4 & 7.2 \\
\bottomrule
\end{tabular}
\label{tc}
\vspace{-15pt}
\end{table}

\noindent\textbf{Qualitative Results. }
% To validate the performance advantage of SFN-YOLO in complex scenarios, Figure~\ref{fig:vis} presents detection results and attention heatmaps for three typical challenging scenes. The model's activation points show clearer and more discrete distributions, with prominent activation centers for each target. The orange areas highlighted regions of the model's attention in each scene, demonstrating its ability to maintain high-precision feature representation and target localization under diverse, challenging environments, thereby outperforming the baseline model.
% in detection performance.
To validate the performance advantage of SFN-YOLO in complex scenarios, Figure~\ref{fig:vis} presents detection results and attention heatmaps for three typical challenging scenes. Compared to the YOLOv8 baseline, SFN-YOLO's activation points show clearer and more discrete distributions, with prominent activation centers for each target. The orange areas highlighted regions of SFN-YOLO's attention in each scene, demonstrating its ability to maintain high-precision feature representation and target localization under diverse, challenging environments.

\begin{itemize}[left=0em]
    \item {Illumination Robustness}: In Figure~\ref{fig:vis}(a), which depicts scenes with strong light-dark contrasts, the global branch constructs illumination-invariant representations using self-attention, generating a channel weight vector that dynamically regulates the local convolution branch. This mechanism suppresses interference from brightness and color channels while enhancing responses from contour and texture channels, enabling precise activation of targets in shadowed areas.
    
    \item {Occlusion Handling}: In Figure~\ref{fig:vis}(b), which shows occlusion scenarios, the global branch forms a pixel-level spatial correlation matrix via cross-attention, leveraging contextual dependencies to reconstruct occluded target features. In contrast to the baseline model, which suffers from fragmented activation due to feature discontinuity, our model generates focused activation, preserving the continuity of occluded target features.
      
    \item {Multiscale Perception}: In Figure~\ref{fig:vis}(c), representing small target scenarios, SFM is embedded at different stages of the backbone network: shallow layers preserve edge details via spatial guidance, while deeper layers incorporate global context to enhance semantic consistency. The hierarchical fusion strategy ensures a balance between pixel-level detail and expansive semantic relationships.

\end{itemize}

The global-local branches form an interactive loop: the global branch enforces semantic priors to filter invalid responses, while the local branch refines localization by correcting spatial biases. This mechanism ensures stable feature representation under challenging conditions, such as illumination shifts, occlusions, and scale variations, leading to a significant performance boost over the baseline.

\noindent\textbf{Out-of-Domain Generalization.} To evaluate the generalization ability of the SFN-YOLO, we tested it on the \textit{Outdoor1} OOD dataset, which includes unseen scenes and conditions. As shown in Table~\ref{t331}, SFN-YOLOn achieves an mAP of 73.2\%, outperforming baselines like YOLOv5n and YOLOv8n. Its top AP75 score of 80.5\% demonstrates strong localization precision, and the leading Average Recall (ARM) of 67.4\% reflects superior object recall. These results validate that our model learns transferable features, making it generalizable for free-range deployment in varying environments.

%\caption{Evaluation on the \textit{Outdoor1} OOD, with the model trained on the ID dataset.}

\begin{table}[t]
\centering
\caption{Evaluation of ID-Trained Model on the \textit{Outdoor1} OOD Benchmark. }
\label{t331}
\renewcommand{\arraystretch}{1} 
\footnotesize 
\begin{tabular}{c|ccc|c} 
\toprule
\textbf{Model} & \textbf{mAP$\uparrow$} & \textbf{AP50$\uparrow$} & \textbf{AP75$\uparrow$} & \textbf{ARM$\uparrow$}\\ 
\midrule
YOLOv5n & \underline{72.8} & 91.1 & \underline{79.2} & 62.0 \\ 
YOLOv8n & 72.6 & 90.7 & 79.1 & 62.2 \\ 
YOLO11n & 72.4 & 91.0 & 78.9 & 61.1 \\ 
RT-DETR-R50 & 69.1 & \textbf{91.3} & 78.1 & \underline{63.8} \\  
MambaYOLO-T & 59.0 & 86.1 & 67.4 & 47.2 \\ 

SFN-YOLOn (Ours) & \textbf{73.2} & \underline{91.2} & \textbf{80.5} & \textbf{67.4} \\ 
\bottomrule
\end{tabular}
\vspace{-15pt}
\end{table}

\begin{table}[t]
\centering
\caption{Ablation Study on Key Components: Global Branch (GB), Local Branch (LB), Channel Guidance (C), and Spatial Guidance (S) on M\mbox{-}SCOPE.}
\label{ab}
\renewcommand{\arraystretch}{1} 
\footnotesize 
\setlength{\tabcolsep}{2.5pt} 
\begin{tabular}{cccc|ccc|cc|cc}
\toprule
\textbf{GB} & \textbf{LB} & \textbf{C} & \textbf{S} & \textbf{mAP$\uparrow$} & \textbf{AP50$\uparrow$} & \textbf{AP75$\uparrow$} & \textbf{APS$\uparrow$} & \textbf{APM$\uparrow$} & \textbf{ARS$\uparrow$} & \textbf{ARM$\uparrow$} \\
\midrule
\xmark & \xmark & \xmark & \xmark & 78.08 & 96.14 & 87.38 & 72.38 & 71.18 & 80.39 & 77.25 \\

\cmark & \xmark & \xmark & \xmark & 77.67 & 96.00 & 87.09 & 71.40 & 70.76 & 80.44 & 77.24 \\

\cmark & \cmark & \xmark & \xmark & 78.19 & 96.14 & 87.38 & 73.24 & 71.10 & 80.89 & 77.17 \\

\cmark & \cmark & \cmark & \xmark & 78.10 & 96.11 & 87.31 & 72.95 & 71.12 & 80.42 & 77.25 \\

\cmark & \cmark & \xmark & \cmark & 78.13 & 96.20 & 87.39 & 73.16 & 71.30 & 80.42 & 76.99 \\

\cmark & \cmark & \cmark & \cmark & \textbf{78.79} & \textbf{96.66} & \textbf{88.86} & \textbf{74.87} & \textbf{73.04} & \textbf{82.40} & \textbf{80.09} \\
\bottomrule
\end{tabular}
\vspace{-15pt}
\end{table}

\noindent\textbf{Ablation Study.} We performed ablation experiments on the M\mbox{-}SCOPE using YOLOv8n as the baseline (Table~\ref{ab}). The baseline mAP is 78.08\%. Adding the Global Branch (GB) reduces it to 77.67\%, while combining it with the Local Branch (GB+LB) raises it to 78.19\%, emphasizing the value of combining global and local features. On the basis of (GB+LB), note that channel guidance (C) or spatial guidance (S) alone may degrade the performance of mAP by introducing bias in feature space. By contrast, using both channel guidance and spatial guidance can significantly improve the performance, indicating that channel guidance and spatial guidance contain complementary information. The full model (GB+LB+C+S) reaches an mAP of 78.79\%, improving the baseline mAP  by 0.71\%, with significant gains in localization (AP75 +1.48\%) and small object detection (APS +2.49\%).

\section{CONCLUSION}
We introduce SFN-YOLO, an efficient real-time detector for poultry. The model fuses local details with global context to significantly enhance object perception amid challenges like scale variation, occlusion, and cluttered backgrounds. Evaluated on our novel M-SCOPE dataset, SFN-YOLO achieves state-of-the-art performance and strong generalization with substantially fewer parameters. This work provides a foundation for automated smart poultry farming.

% \vfill\pagebreak
\bibliographystyle{IEEEbib}
\bibliography{strings,refs}

\end{document}